\documentclass{article}

\usepackage{microtype}
\usepackage{graphicx}
\usepackage{subcaption}
\usepackage{booktabs} 
\usepackage{xcolor}
\usepackage[most]{tcolorbox}
 \usepackage{svg} 
\usepackage{booktabs}
\usepackage{tabularx}
\usepackage{array}
\usepackage{url} 
\usepackage{hyperref}

\usepackage[preprint]{icml2026}

\usepackage{amsmath}
\usepackage{amssymb}
\usepackage{mathtools}
\usepackage{amsthm}

\usepackage[capitalize,noabbrev]{cleveref}

\theoremstyle{plain}

\theoremstyle{definition}

\theoremstyle{remark}

\usepackage[textsize=tiny]{todonotes}

\icmltitlerunning{AI Must Embrace Specialization via Superhuman Adaptable Intelligence}

\begin{document}

\twocolumn[
  \icmltitle{AI Must Embrace Specialization via Superhuman Adaptable Intelligence}



  \icmlsetsymbol{equal}{*}
\begin{icmlauthorlist}
  \icmlauthor{Judah Goldfeder}{equal,col}
  \icmlauthor{Philippe Wyder}{equal,distyl}
  \icmlauthor{Yann LeCun}{nyu}
  \icmlauthor{Ravid Shwartz-Ziv}{nyu}
\end{icmlauthorlist}

\icmlaffiliation{col}{Columbia University, New York, NY, USA}
\icmlaffiliation{distyl}{Distyl, New York, NY, USA}
\icmlaffiliation{nyu}{New York University, New York, NY, USA}

\icmlcorrespondingauthor{Judah Goldfeder}{jag2396@columbia.edu}

  \icmlkeywords{Machine Learning, ICML}

  \vskip 0.3in
]



\printAffiliationsAndNotice{}  

\begin{abstract}
Everyone from AI executives and researchers to doomsayers, politicians, and activists is talking about Artificial General Intelligence (AGI). Yet, they often don’t seem to agree on its exact definition. One common definition of AGI is an AI that can do everything a human can do, but are humans truly general? In this paper, we address what’s wrong with our conception of AGI, and why, even in its most coherent formulation, it is a flawed concept to describe the future of AI. We explore whether the most widely accepted definitions are plausible, useful, and truly general.
We argue that AI must embrace specialization, rather than strive for generality, and in its specialization strive for superhuman performance, and introduce Superhuman Adaptable Intelligence (SAI). SAI is defined as intelligence that can learn to exceed humans at anything important that we can do, and that can fill in the skill gaps where humans are incapable. We then lay out how SAI can help hone a discussion around AI that was blurred by an overloaded definition of AGI, and extrapolate the implications of using it as a guide for the future.
\end{abstract}

\section{Introduction}
The AI community has become increasingly fractured over where the field is headed. On one side are “doomers,” who argue we are headed towards a gruesome societal endgame---mass unemployment, loss of human agency, and a future in which humanity becomes subordinate to artificial overlords. On the other side are those who expect advanced artificial intelligence to bring something close to utopia, ending hunger, suffering, and scarcity. A third camp frames AI as a “normal technology,” forecasting major impacts but rejecting extreme narratives~\cite{narayanan2025ai}.

Central to all of these views is the concept of Artificial General Intelligence or AGI. Yet, as is often the case in widely public debates, much of the disagreement stems less from evidence than from terminology: AGI is invoked constantly, but rarely defined precisely, and the resulting ambiguity has made the debate far more confusing and far more polarized—than it needs to be. 

Much of the discourse uses human intelligence as a paradigm of generality, but we argue that this notion is fundamentally misguided. As humans, we struggle to perceive our own blind spots; this leads to the illusion of generality. In truth, we are only good at the specific subset of tasks that are important to our existence, but are completely incapable of performing tasks outside this narrow range.
Awareness of human limitation gives rise to a critical realization: humans may be specialized creatures, but are nonetheless capable of accomplishing or quickly learning a wide range of incredible things. We argue that the current focus on AGI and generality as the North Star of the field, should be replaced with an emphasis on adaptability, including the time it takes to learn a new task, and the range of tasks capable of being learned. We refer to this as \textbf{Superhuman Adaptable Intelligence (SAI)}.
A natural corollary of an emphasis on adaptability is the need for a model with strong assumptions about the world. This suggests \textbf{self-supervised learning (SSL)} as a promising way to acquire generic knowledge, and \textbf{world models} as a useful mechanism for planning and zero-shot task transfer.
We believe that recentering the discourse around SAI will lead towards better communication, clearer goals, and more rapid progress.




\begin{tcolorbox}[
enhanced,
colback=gray!3,
colframe=black!60,
arc=2mm,
boxrule=0.6pt,
left=6pt,right=6pt,top=6pt,bottom=6pt
]
\setlength{\parskip}{4pt}

{\color{teal!70!black}\textbf{[Pos.\ \#1]} Human intelligence is not general in any meaningful way}

{\color{blue!70!black}\textbf{[Pos.\ \#2]} Generality is not a requirement for an intelligence to be extremely useful}

{\color{violet!75!black}\textbf{[Pos.\ \#3]} There is no consensus on the meaning of the term AGI in industry or academia}

{\color{orange!85!black}\textbf{[Pos.\ \#4]} Existing definitions are insufficient}

{\color{red!70!black}\textbf{[Pos.\ \#5]} We should instead focus on Superhuman Adaptable Intelligence, which points toward SSL and world models}

\end{tcolorbox} 

\section{Human Intelligence is Specialized}\label{sec_HumanIntelIsSpecialized}

While the idea of human intelligence as the paradigm of generality is ubiquitous in the literature, two related but distinct notions of generality are often conflated:

\begin{enumerate}
  \item The average, educated human is capable of a wide range of tasks that are very ``general`` in nature, and enable a wide range of objectives to be accomplished. This includes things like complex planning and locomotion, fine motor skills, abstract thinking, self simulation, spatial reasoning, and visual understanding.
  \item Human intelligence as a whole is ``general`` because it can be specialized to ``any`` given task, whether it be medicine, advanced mathematics, plumbing, or playing chess.
\end{enumerate}

Both of these claims make the same error: \textbf{circularly defining generality in human terms, and then asserting humanity as its paradigm.}

Evolution has honed humanity over time to be highly specialized in the domain of skills necessary for survival in the physical world. The things most innate to us are not always the most simple, but the most critical for our survival. This observation has given rise to Moravec's Paradox, where the tasks we find easiest, like locomotion, are difficult for computers, but tasks that we find difficult, which are not essential to our survival, like playing chess, turn out to be much easier for computers. This clearly illustrates the illusion of our generality. While the average abilities of an educated human are truly remarkable, one only need ask them to play chess like a grandmaster, or compose a musical symphony like Beethoven, to truly realize the hubris of calling such an intelligence general.

The above argument serves to dispel the first definition of generality. The second notion of generality is more subtle in its error. By identifying specialization/adaptation as the core component of generality, it is closer to the definition of SAI that we are arguing in favor of. However, our point of contention is that we object to calling human adaptation general. While we are excellent at adapting to the tasks that were of high evolutionary importance, we are simply incapable of adapting to many tasks outside of this range at a high level. Take chess as an example. Magnus Carlsen is widely regarded as the greatest chess player of all time, and as such represents the pinnacle of human adaptation when it comes to playing chess. But this begs the question: Is Magnus actually any good at chess? When compared with the best computers, the answer is clearly no. Even more damning is that with modern day computers, creating a program that plays chess at a much higher level than Magnus \textit{is not particularly difficult}. Our perception of his ability is colored by the limitations of humanity. Humans in general are bad at chess; Magnus is much better than most humans. The conclusion, that Magnus is good at chess, is a perfect illustration of our own human centric biases.  Magnus Carlsen is not objectively good at chess, he is good at chess with respect to human performance levels. By any objective metric, playing chess at a much higher level is not  difficult from a computational perspective, but it is something that humans are incapable of. Relatedly, many animals can perform tasks that humans cannot do at a high level, such as echolocation.

So what are humans then, if not a paradigm of general intelligence? The evidence points to specialized adaptation. We have an incredible ability to adapt and specialize, within the range of tasks that we evolved to address~\cite{russell_norvig_2010_AIModernApproach}. This is the main contention of \textcolor{teal!70!black}{\textbf{[Pos.\ \#1]}}.

\subsection*{Alternative Views}
Several objections have been raised to our assertion that humans are not general. Elon Musk and Demis Hassabis have claimed that our argument conflates General Intelligence with Universal Intelligence. They further argue that the human brain is indeed general in the Turing Machine sense, capable of learning anything computable given enough time, memory, and data. They therefore claim that  "brains are the most exquisite and complex phenomena we know of in the universe (so far), and they are in fact extremely general"~\cite{Hassabis2025XUniversalIntelligence,Musk2025DemisIsRight}.

In response, it is indeed important to clarify terms. Universal Intelligence refers to the ability to act intelligently over all computable environments~\cite{legg2007universal}. General Intelligence, as Demis is using it, seemingly refers to adaptation to any computable task given time and resources. Far from conflating the terms, we are  arguing that humans are not capable of either of these things. 

While the issue of whether approximate Turing-completeness under idealized conditions matters for defining intelligence is a legitimate question, it is missing the point.
Even if we grant the fact that human brains are approximately Turing-complete (far from an obvious fact), under real constraints such as finite memory, finite time, and finite attention, we handle only a tiny sliver of possible problems. 
The space of possible functions is unimaginably vast, and we can represent an infinitesimal fraction. We feel general because we can't perceive our blind spots, not because we lack them.

\section{Implications for AI North Star Terminology}

\begin{figure*}[ht]
  \vskip 0.2in
  \begin{center}
    \centerline{\includegraphics[width=\textwidth]{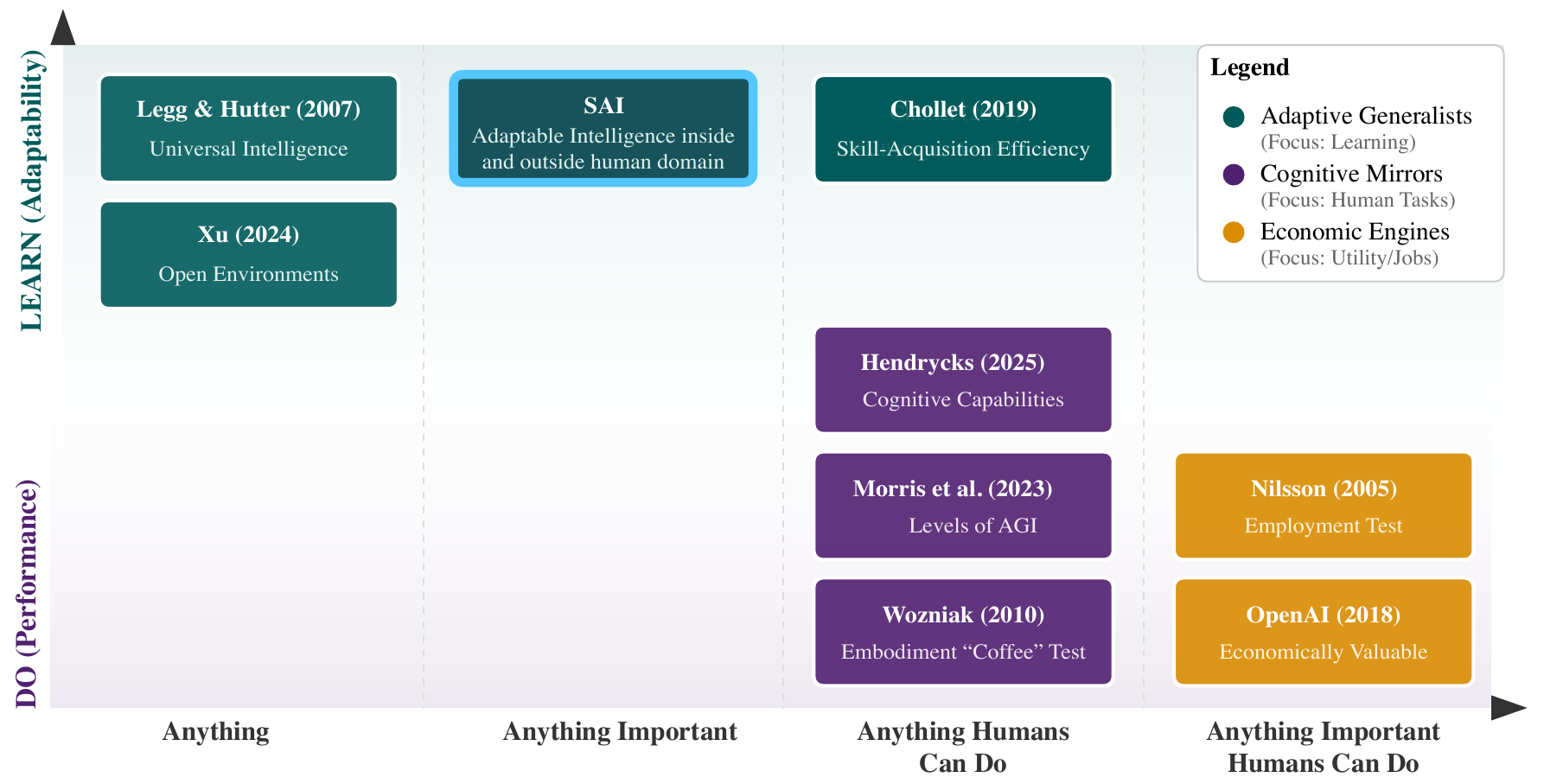}}
    \caption{A two-dimensional semantic map organizing prominent definitions for AGI and other North Star measures of artificial intelligence, along two axes. The vertical axis represents the source of intelligence, ranging from performance-based capabilities (DO, bottom) to learning and adaptability (LEARN, top). The horizontal axis represents the scope of tasks, from universal/open-ended domains (left) to human-centric and economically-focused domains (right). Definitions cluster into three categories: Adaptive Generalists (teal) emphasize learning efficiency and generalization in open environments; Cognitive Mirrors (violet) focus on replicating human-level cognitive capabilities across broad task domains; Economic Engines (orange) prioritize practical utility and economic value in human-relevant tasks. \textit{Superhuman Adaptable Intelligence (SAI)} falls into the realm of adaptable AI that can do anything that is important both inside and outside the human realm.
    }
    \label{fig_axesOfAGI}
  \end{center}
\end{figure*}

\subsection{A Survey of Definitions}
Measuring machine intelligence is non-trivial. Language-based tests, such as the Turing Test, where a machine has to pretend to be a human well enough to fool a human to believe the machine is human \cite{turing1950_TuringTest} and the Winograd schema challenge that tests common-sense reasoning and natural language understanding \cite{Levesque_WinogradSchemaChallenge} are helpful to measure aspects of intelligence, but not a true measure of whether AGI is achieved. Steve Wozniak's \textit{Coffee Test}---whether a machine could make a cup of coffee if sent to a random kitchen---draws attention to the fact that, despite claims to the contrary\cite{chen_belkin_bergen_danks_2026}, language alone is not sufficient to be considered  intelligent and that human intelligence \textit{adapts} well to unseen environments~\cite{Wozniak2010_CoffeeTest}. 
AGI definitions commonly fall into categories along two axes, the first one defining what capabilities we are referring to, and the second defining the required scope of those capabilities:
\begin{enumerate}
    \item Axis 1 (capability): (A) AI that can \textbf{learn} to do tasks vs (B) AI that can \textbf{do} tasks out of the box
    \item Axis 2 (scope): (I) Anything, (II) Anything important, (III) Anything humans can do, (IV) Anything humans can do that is important
\end{enumerate}
We visualized popular definitions of AGI in accordance with this two-dimensional framework in Fig. \ref{fig_axesOfAGI}.

There is a reasonable argument to be made for a third axis that spans the space from observable capability to subjective understanding~\cite{Searle_1980_sentience}, thereby including the dimension of the \textit{internalist} view. According to this view, an AI could meet any performance benchmark for AGI, yet if it lacks subjective experience (qualia), it remains merely a simulation of intelligence rather than the genuine article. While the exploration of this dimension is profound, we consider it outside the scope of this work. Our focus is on operational definitions—metrics that can be observed and measured—whereas the internalist objection currently resides in the realm of metaphysics and philosophy of mind, offering no falsifiable test for engineering progress.

Regardless, one thing is clear: AGI as a term is overloaded with varying definitions from high-impact sources. This confusion has even led to claims that AGI has already arrived\cite{aguerra_y_arcas_norvig_2023,chen_belkin_bergen_danks_2026}. The varying definitions plotted in Fig. \ref{fig_axesOfAGI} and the imprecise nature of the public discussions being had by high-profile individuals around AGI, as shown in the previous section, clearly demonstrate \textcolor{violet!75!black}{\textbf{[Pos.\ \#3]}}.

\subsection{Why Existing Definitions are Insufficient}

\textcolor{teal!70!black}{\textbf{[Pos.\ \#1]}} has the following implications for these definitions:

\begin{enumerate}
  \item Humanity is still \textit{quite useful}, so AI does not need to be general to still be groundbreaking and powerful (\textcolor{blue!70!black}{\textbf{[Pos.\ \#2]}}).
  \item Any definition focused exclusively on humanity as a goal cannot claim to be general.
  \item Focusing exclusively on humans is also not ideal, since there are many tasks we cannot do that are still high utility and important.
\end{enumerate}

In addition, for a definition to be useful, it must meet the following criteria:

\begin{enumerate}
  \item It must be feasible. If a goal is not possible to be realized from a theoretical perspective, it provides questionable value.
  \item It must be internally consistent. If a definition claims to be \textit{general}, it must actually \textit{be general} in a meaningful way.
  \item It must be assessable. The goal as presented should lead to clear subgoals and strategies, and there must be a clear metric with which to measure progress.
\end{enumerate}

Having established these criteria, we can now demonstrate \textcolor{orange!85!black}{\textbf{[Pos.\ \#4]}}, namely that existing definitions come up short.

\begin{table*}[t]
\centering
\small
\setlength{\tabcolsep}{6pt}
\renewcommand{\arraystretch}{1.25}

\begin{tabularx}{\linewidth}{%
  >{\raggedright\arraybackslash}X
  >{\raggedright\arraybackslash}p{0.11\linewidth}
  >{\raggedright\arraybackslash}p{0.39\linewidth}
}
\toprule
\textbf{AGI Definition} & \textbf{Failure mode} & \textbf{Explanation} \\
\midrule

\textit{``Match or exceed the cognitive versatility and proficiency of a well-educated adult.''} \cite{hendrycks2025definitionagi}
& Not Consistent & Human cognition is not general in any meaningful way. This definition is also unnecessarily narrow\\

\textit{``Highly autonomous systems that \textbf{outperform humans at most economically valuable work}.'' \cite{Morris_LevelsOfAGI}}
& Not Consistent  & The focus here is explicitly  on a subset of tasks that are of economic worth. Clearly not General\\

\textit{``A system that should be able to do \textbf{pretty much any cognitive task that humans can do}.'' --- \textbf{Demis Hassabis (DeepMind CEO)} \cite{Mitchell2024_DebatesOnTheNatureOfAGI}}
& Not Consistent & This definition is  not actually general both in its focus on humans, and also  in its focus on "cognitive tasks", which seems to be to the exclusion of physical tasks like locomotion \\

``We need precise, quantitative definitions and measures of intelligence – in particular human-like general intelligence....
...The intelligence of a system is a measure of its \textbf{skill-acquisition efficiency} over a scope of tasks, with respect to priors, experience, and generalization difficulty'' \cite{chollet2019measureintelligence}
& Not Consistent & Chollet himself admits as much, calling human cognition 'only “general” in a limited sense', a contradiction of terms \\

\textit{``We define AGI as a system that demonstrates \textbf{broad generality} (performing a wide range of tasks) and \textbf{high performance} (matching or exceeding human levels).'' \cite{Morris_LevelsOfAGI}}
& Not Feasible & While they acknowledge generality requires exceeding human levels, with a focus on direct performance over adaptation, such a system is not realizable with finite resources. \\

``Intelligence measures an agent’s ability to \textbf{achieve goals in a wide range of environments}.'' \cite{legg2007universal}
& Not Feasible & Legg and Hutter define the domain of environments as all that are computable. They further emphasize ability over adaptability. Strong ability on such a vast set of tasks is not realizable with finite resources\\

"Highly autonomous systems that outperform humans at most economically valuable work" \cite{openai_charter}. & Not Assessable & The focus on performance means that any evaluation would have to benchmark against an ever growing set of tasks\\ 

\bottomrule
\end{tabularx}
\caption{The failure of most AGI definitions. Note: some definitions fail for multiple reasons, but we only highlight one.}
\label{tab:definitions}
\end{table*}


First, definitions of AGI that claim true generality fall prey to the "No Free Lunch" theorem---no single, general-purpose machine learning algorithm or optimization strategy works best for every problem~\cite{Wolpert_NoFreeLunch}. Or to frame it differently: given finite energy, an approach that directs available energy towards learning a finite set of tasks will reasonably outperform an approach that distributed the finite energy over an infinite amount of tasks. At the limit, the amount of energy dedicated to each of the infinite tasks approaches zero. Thus, any definition that defines the scope as \textit{literally anything computable} fails our criteria by \textit{not being feasible}.

Second, any definition of AGI that focuses on a subset of tasks, or that emphasizes specialization and adaptation as key metrics, can not truly be said to be general. Similarly, AGI measured by the "general" nature of humans is not truly general. Chollet acknowledges this problem and states that human intelligence "is only “general” in a limited sense"~\cite{chollet2019measureintelligence}, but we contend that this is simply an inherent contradiction of terminology. For this reason, Shane Legg and Marcus Hutter speak of "Universal Intelligence" rather than AGI because human intelligence is "far too limited"~\cite{legg2007universal}, since a definition of AGI that is human-centric excludes the infinite space of non-human intelligence~\cite{Wang2019_OnDefiningArtificialIntelligence}. 
Despite the above objections, AGI defined specifically as the ability to match human cognitive breadth is quite popular: Hendrycks et al. and Morris et al. argue that human generality is the only general example for the concept of intelligence~\cite{hendrycks2025definitionagi, Morris_LevelsOfAGI}. 
While the above emphasized the ability to do \textit{anything} humans can do, others
argue that AGI must be able to learn to do or do anything \textit{important} that humans can do. Their arguments acknowledge that the domain of human intelligence is finite and that it is desirable for AI to be able to perform or learn to perform a subset of important tasks: tasks that generate economic value. In the words of Nilsson, "Systems with true human-level intelligence should be able to perform the tasks for which humans get paid" ~\cite{Nilsson_EmploymentTest}.\footnote{Nilsson doesn't use the term AGI, he speaks of "strong AI" or "human-level artificial intelligence" (the term was popularized later by Shane Legg and Ben Goertzel), but still pushes the idea of humans being "more-or-less" general purpose}, an idea further echoed in the OpenAI Charter~\cite{openai_charter}.

We suspect that the cause for such a widespread conflation of human intelligence with generality stems from the urge for self-flattery, and the difficulty of truly conceiving of our own limitations. Regardless, all the definitions in this category fail our criteria by \textit{not being internally consistent}.

One might raise the objection that our contention here is merely one of semantics, and that these definitions can still be valuable North Stars for the field, even if they misuse the term "general". In response, we argue that when defining the end goal of an entire field, semantics are extremely important. A misapprehension of generality is dangerous for several reasons. It obscures how such an intelligence can actually be realized, which violates our feasibility criteria. Further, it can lead to unnecessarily narrow conceptions of what the end goal should be. For example, the belief that humans are general has led to several definitions of AGI as mimicking humans, which is certainly far too limited a goal for what AI is and can become.

Third, definitions of AGI that cannot be assessed or evaluated are not practical or useful. Legg and Hutter acknowledge this issue in their paper: "various practical challenges will need to be addressed before universal intelligence can be used to construct an effective intelligence test"~\cite{legg2007universal}. The ability to measure progress is critical for many reasons. An enormous body of evidence suggests that the precise ability to measure progress is one of the strongest catalysts of \textit{progress itself}~\cite{wyder2025common}. Relatedly, clear metrics usually give an idea of what sorts of subgoals and strategies are useful. Even more fundamentally, a definition that is not measurable is not really much of a definition at all, and is often indicative of a lack of precision, or a hand-wavy nature.

This criterion highlights a key difference between the two categories of definitions on our first axis (capability). Any definition that focuses on \textit{learning} or \textit{adapting} implicitly has a clear metric with which to evaluate intelligence: speed of adaptation to new tasks. Conversely, definitions focused on \textit{doing} and \textit{performing} often lack any obvious way to measure this, other than benchmarking the AI's ability to do everything, an ever-expanding and ill-defined set of benchmarks.
Table \ref{tab:definitions} elaborates on our issues with many popular AGI definitions (\textcolor{orange!85!black}{\textbf{[Pos.\ \#4]}}).

\section{Why Specialization Wins}
To motivate \textcolor{red!70!black}{\textbf{[Pos.\ \#5]}}, it behooves us to explore the importance of specialization. Specialization is not an accident of biology; it is a predictable consequence of limited resources, competing objectives, and environments that reward performance on a small subset of evolutionarily relevant challenges. Forister et al. state that a generalist organism carries genetic traits suited to various environments, but never the ideal combination for thriving in any one of them~\cite{Forister_RevisitingEvoOfEcoSpec}. Organisms face persistent trade-offs: improving performance on one niche often reduces performance elsewhere, and selection therefore tends to favor designs that are sharply tuned to the local payoff landscape rather than uniformly competent across all possible conditions~\cite{Futuyma1988_EvoOfEcoSpec}. In markets and organizations, the same logic appears under a different name: entities that fail to meet the performance threshold disappear, so competition acts as a selection mechanism that amplifies effective strategies and eliminates ineffective ones ~\cite{Hannan_Freeman_1977_EcologyOfOrganizations,Loasby_1983_EvoTheoEconChange}. AI systems are not exempt from this pressure: models that are too costly, too unreliable, or insufficiently accurate in the domains that matter will be neglected in favor of systems that are better matched to those domains.

In machine learning, the core mathematical point is that performance gains require assumptions about the problem class i.e. the target distribution. Again, “No Free Lunch”. An algorithm wins by being a good fit for the target problem. As AI improves, specialized systems can improve too: if it is possible to attain a higher performance on a task, a system that concentrates that capability on a narrower task can typically realize larger gains than a system that must spend capacity and compute covering additional unrelated tasks.

Practically, this means that generality is intractable. Although multi-task learning can benefit performance when tasks share an underlying structure, it can lead to "negative transfer" when tasks compete for representational capacity or impose conflicting gradients, and thereby harm task performance~\cite{ruder2017overviewmultitasklearningdeep}. Models that route queries to specialized subsets of model parameters depending on the task are a technological acknowledgment of this limitation---these systems attain breadth and scale through repeated, modular specialization rather than uniform shared parameters for all inputs\cite{Fedus2022_MixtureOfExperts}. Although seemingly "general", these models achieve their best performance through internal specialization.


Universal generality is a theoretical concept, but in practical terms it is a myth. A large fraction of what we intuitively mean by ``do anything'' reduces to planning and decision-making under uncertainty. Classical planning problems quickly become intractable in worst case (e.g., propositional STRIPS variants)~\cite{Bylander1994}, and probabilistic planning inherits similarly severe complexity barriers~\cite{LittmanGoldsmithMundhenk1998}. This does not mean planning is impossible in practice; it means that broad generality across arbitrary environments has no reason to be computationally cheap. A specialized agent that restricts the space of environments, goals, and action models it must handle can leverage structure and avoid worst-case blowups. This is similar for humans, as our biases, genetic makeup, and environment naturally drive us towards “human things,” a mere sliver of universal generality.  

Empirically, specialized AI systems repeatedly demonstrate the advantage of concentrating model design, data curation, and evaluation around a single domain objective. Protein structure prediction is an archetypal example: AlphaFold achieved dramatic gains by targeting a specific scientific task with task-specific training and architectural choices, and it set a new bar for accuracy and usefulness in that domain~\cite{Jumper2021}. It is therefore plausible---indeed, expected under both the No Free Lunch framing and negative-transfer dynamics---that an AI system asked to ``fold proteins \emph{and} fold laundry'' will not match a protein-folding specialist on protein-folding performance unless it internally recovers specialization (e.g., via routing, modularity, or dedicated submodels)~\cite{Wolpert_NoFreeLunch,ruder2017overviewmultitasklearningdeep,Fedus2022_MixtureOfExperts,Jumper2021}.

Specialization also clarifies why AI can be uniquely valuable: it can target precisely the domains where human cognition is systematically miscalibrated. Humans exhibit stable biases and heuristics that are often sensible under ancestral constraints but error-prone in modern settings~\cite{TverskyKahneman1974}. More broadly, the evolutionary mismatch hypothesis argues that many psychological mechanisms were tuned for past selection regimes and can therefore produce maladaptive outputs in contemporary environments~\cite{Li2018Mismatch}. This creates an opportunity: specialized AI systems can be designed to excel exactly where humans are weak but where correctness now matters (e.g., high-dimensional statistical inference, optimization under constraints, complex mechanistic modeling)~\cite{TverskyKahneman1974,Domingos2012}.

Finally, none of this implies that generality is “bad”. It implies a narrower, more operational claim: we must embrace specialization rather than fight it. Even in domains that feel like demonstrations of ``general intelligence,'' the history of AI milestones frequently reflects intense domain targeting rather than broad competence, while newer ``general'' methods still succeed by exploiting strong structure in the task family~\cite{Silver2018_AlphaGo}. For high-stakes applications (e.g., scientific discovery, medicine), the correct aspiration is not to preserve the romance of a single generalist mind, but to build the strongest available specialists---and, where needed, compose them into systems whose coordination is engineered rather than assumed.

We should also note that this claim does not dispute the bitter lesson~\cite{sutton2019bitter}. The bitter lesson is the observation that approaches that scale with computational power tend to outperform ones based on domain knowledge, a claim that we agree with. The diminishing usefulness of domain knowledge is distinct from the usefulness of domain specialization. As scaling progresses, we will need to know less about proteins to build a system that does protein folding; however, such a system still benefits from focusing \textit{specifically} on proteins.

\begin{figure}[ht]
  \centering
  \includegraphics[width=1.0\linewidth]{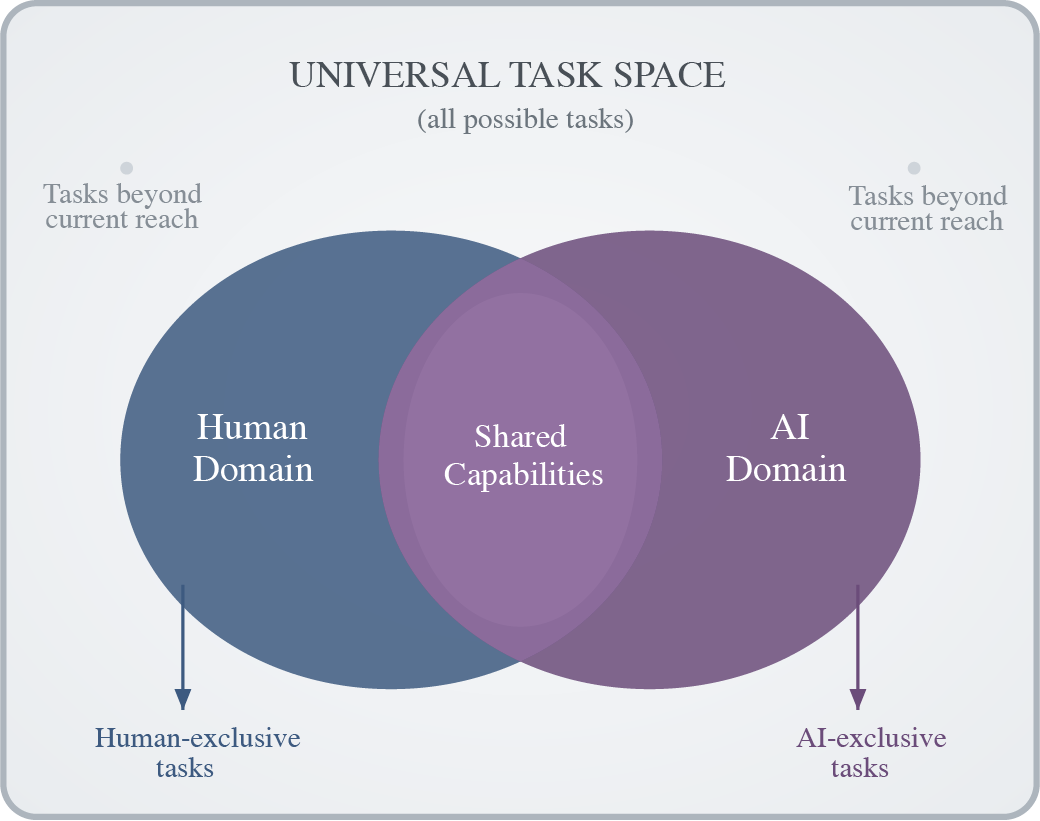}
  \caption{Illustration of the task space overlap between the human domain and the AI domain within the universal task space.}
  \label{fig_Uni_AI_Hum_domain}
\end{figure}
Awareness of the "narrowness" of humans and the benefit of specialization allows us to exploit the complementary nature of AI as it is filling in the gaps in the human domain where it matches or eclipses human performance, while also being able to perform tasks outside the human domain (see figure \ref{fig_Uni_AI_Hum_domain}).

\section{Towards Superhuman Adaptable Intelligence}
Given the utility of specialization, we propose \textbf{Superhuman Adaptable Intelligence (SAI)} as the \emph{idée fixe} of AI research (\textcolor{red!70!black}{\textbf{[Pos.\ \#5]}}). Unlike the earlier AI North Star terminologies that we challenged, our definition of SAI sidesteps issues of feasibility by focusing on adaptation to tasks with human utility, as opposed to the performance of simply doing the task. We embrace the necessity of specialization, and avoid the pitfall of claiming generality. Further, we broaden the task domain beyond the human task domain, while not requiring that the AI is master of the human task domain as a whole. Finally, adaptation speed---the speed with which an agent can acquire new skills and learn new tasks, can be measured, and thus our approach is practical. 

\begin{tcolorbox}[
  enhanced,
  colback=blue!2,
  colframe=blue!55!black,
  arc=2mm,
  boxrule=0.8pt,
  left=8pt,right=8pt,top=8pt,bottom=8pt
]
{\color{blue!70!black}\large\bfseries Definition}\par
\vspace{4pt}
\noindent
\textbf{Superhuman Adaptable Intelligence (SAI)} is capable of adapting to exceed humans at any task humans can do, while also being able to adapt to tasks outside the human domain that have utility.
\end{tcolorbox}

Our definition is most similar to Chollet's~\cite{chollet2019measureintelligence}, except that we object to calling such a definition general, and also reject his view that we "should benchmark progress specifically against human intelligence".
While human performance can be a useful reference point during early development, we argue that anchoring benchmarks to human baselines is ultimately orthogonal to the route to superhuman capability. AI models and systems that optimize well-defined objectives and improve through self-play, evolutionary search, or large-scale exploration in simulation can surpass human performance without imitation~\cite{zhao2025absolutezeroreinforcedselfplay}. We believe that over-indexing on ``human-level'' metrics risks misspecifying the target and limiting evaluation to anthropocentric tasks and constraints.

More broadly, any evaluation scheme that treats intelligence as a checklist of fixed competencies---whether anchored to human baselines or to an ever-growing catalog of tasks---misses the point of SAI. Instead, the focus should be on minimizing adaptation time. The space of possible skills is effectively unbounded, so individually testing skills becomes a Sisyphean endeavor.

\begin{tcolorbox}[
  enhanced,
  colback=teal!2,
  colframe=teal!55!black,
  arc=2mm,
  boxrule=0.8pt,
  left=8pt,right=8pt,top=8pt,bottom=8pt
]
{\color{teal!70!black}\large\bfseries Metric}\par
\vspace{4pt}
\noindent
\textbf{SAI} is measured by the \textbf{speed} with which it takes an agent to acquire new skills and learn new tasks.
\end{tcolorbox}

Our vision towards SAI as a North Star is potentially realizable via self-supervised learning (SSL). We believe that learning in the embedding space as opposed to in the token space may drive performance gains. We also believe that world models may help us advance towards SAI. Simultaneously, we reject the concept of a single model or architecture as the "one paradigm to rule them all,"  as it would suggest that the evolution of artificial intelligence will come to a halt once that architecture has been discovered.

It is also important to note that our definition emphasized tasks outside the human domain that have \textit{utility}. The purpose of this clause was to exclude a potential infinitely set of useless tasks from our definition, but we have not as of yet precisely defined utility, or how we determine task importance. Many definitions have been proposed, such as economic value or societal agreement. The exact definition one prefers is largely orthogonal to our arguments here, and we leave debating which one is most appropriate to other work.

\begin{figure}[t]
  \centering
  \includegraphics[width=0.9\linewidth]{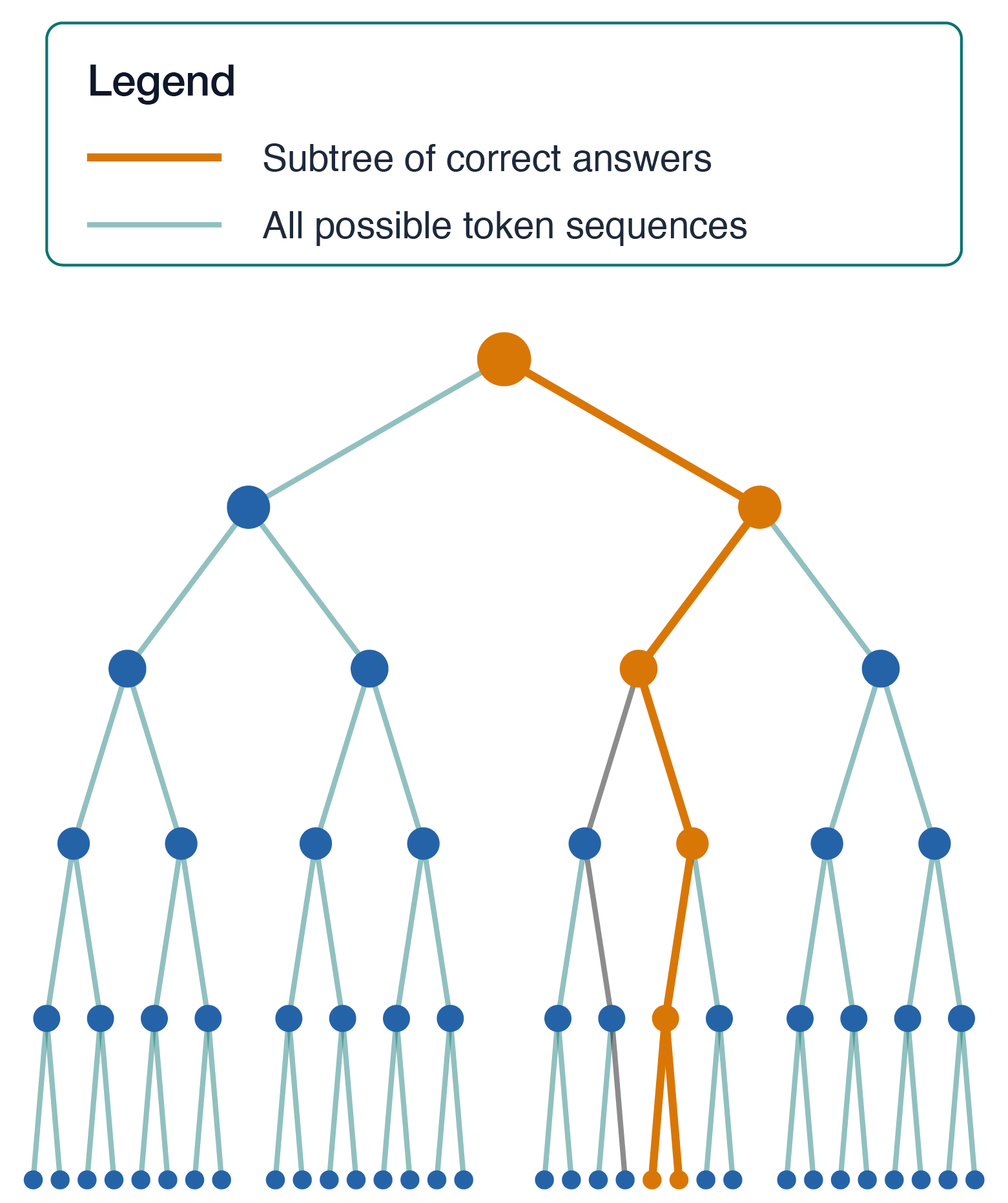}
  \caption{Illustration of autoregressive model divergence}
  \label{fig_divergence}
\end{figure}

\subsection{Why Self Supervised Learning}
By shifting the focus from performance to adaptation, SAI points to SSL as a potential pathway. Specializing to a wide range of tasks requires the ability to learn generic knowledge. In many real-world settings, supervised learning is not feasible in practice because it presupposes access to large, reliably labeled datasets~\cite{LeCun2015DeepLearning}---an assumption that often fails outside carefully curated benchmarks. In contrast, SSL can be applied to any data that contain exploitable internal structure~\cite{balestriero2023cookbookselfsupervisedlearning}. Further, perhaps even more powerful, SSL has actually been shown to be on par with and even exceed SL \textit{even when supervision is abundant}~\cite{he2020momentum,grill2020bootstrap,chen2020simple,he2022masked}. SSL fueled the rise of GPTs, and has reached SOTA performance in most domains.

\subsection{The substrate for fast adaptation}
Adaptation and specialization can be produced by many architectures and paradigms, yet which architecture is most performant remains an open research question. Designing maximally adaptable algorithms remains a central pursuit of meta learning~\cite{finn2017model}. The brain is not a monolith, but a system of systems. This suggests that no single system will be able to adapt in the way that humans do. Thus, we believe that adaptation requires hierarchy and diversity of models and modalities.

Specifically, we believe that adaptation is benefitted by a world model and by moving from token level prediction to latent prediction architectures such as Dreamer 4, Genie 2, or Joint Embedding Prediction Architecture  (JEPA)~\cite{van2025joint, Assran2023_I_JEPA, hafner2025_Dreamer4, bruce2024_genie_generativeinteractiveenvironments}. Pixels are not state. The physical world is too rich and too stochastic for pixel-level prediction to be a meaningful objective; what matters is learning and forecasting a compact representation that captures the system's dynamics.
It has long been posited that humans and animals make heavy use of world models in their cognition~\cite{craik1967nature}. A world model allows for simulation, and therefore planning~\cite{Schrittwieser2020}. As such, it is the hallmark of zero shot and few shot adaptation~\cite{lecun2022path}. Although, we find this argument towards a particular group of architectures persuasive, SAI doesn't dictate a specific architecture.

\subsection{On the importance of diversity}
Homogeneity kills research.
Autoregressive LLMs and LMMs have become the dominant architecture in the state-of-the-art "general" AI space~\cite{Huang2024_LLM2LMM, su2025largelanguagemodelsreally}. This concentration is understandable---shared tooling and benchmarks create momentum---but it also narrows the search space. Progress is most rapid when a greater diversity of solutions are explored.

In addition to slowing progress, these homogeneous solutions are often only local optima. GPTs and similar autoregressive models are no exception, they have many flaws~\cite{lin2021limitations}; their errors diverge exponentially with prediction length~\cite{lecun_objective_driven_ai_harvard_2024}, as shown in figure \ref{fig_divergence}. In practice, compounding prediction error makes long-horizon interaction brittle.
SAI counters homogenization and drives diversity in AI development. It provides a more coherent and reasonable target that fosters a diversity of specialization profiles. Embracing specialization  counteracts incentives that lead to fast convergence towards the mean.

\section{Discussion}

The AGI discourse is often framed as a single destination, benchmarked against an ill-defined notion of ``human-level'' generality. We argue that this framing is both scientifically unhelpful and operationally misleading. Human intelligence is not a universal competence engine; it's a collection of specialized capabilities shaped by constraints and selective pressures. There is no reason to expect the most capable artificial systems to mirror the human task distribution, nor to treat human performance as the natural reference point for progress.

We propose \textit{Superhuman Adaptable Intelligence (SAI)} as a more concrete and productive North Star: the ability to rapidly adapt to \emph{important} tasks inside and outside the human domain. The central quantity is not a checklist of skills, but the speed and efficiency with which new skills are acquired under realistic resource constraints. This reframes evaluation away from human-centric benchmarks and toward measurable adaptation dynamics.

\begin{tcolorbox}[
  enhanced,
  colback=red!2,
  colframe=red!55!black,
  arc=2mm,
  boxrule=0.8pt,
  left=8pt,right=8pt,top=8pt,bottom=8pt
]
{\color{red!70!black}\large\bfseries Key Insight}\par
\vspace{4pt}
\noindent
The AI that folds our proteins should not be the AI that folds our laundry!
\end{tcolorbox}

Finally, SAI's specialization focus fosters an environment that promotes diverse engineering approaches. Progress won't come from a single architecture optimized for next-token prediction. We believe instead that systems that learn general latent structure from unlabeled data, build world models that support planning, and compose specialized modules are better suited to fast adaptation. 
Put another way: it is highly unlikely that an AI tasked to fold both proteins and laundry will exceed a protein-folding specialist at protein folding or a laundry-folding specialist at laundry folding. Given limited resources, capability should be allocated to the tasks that carry utility rather than to an anthropocentric notion of universal competence. 
One promising path forward is therefore to emphasize self-supervised learning approaches, predictive world models, and modularity---and to judge advances by how quickly and reliably they produce new competence, rather than by how closely they imitate human behavior.

\bibliography{bibliography}

@misc{Wozniak2010_CoffeeTest,
  author       = {Wozniak, Steve},
  title        = {Could a Computer Make a Cup of Coffee?},
  howpublished = {Fast Company Live},
  year         = {2010},
  month        = {March},
  url          = {https://www.fastcompany.com},
  note         = {Interview proposing a physical benchmark for AGI},
  urldate = {2024-05-20}
}

@article{turing1950_TuringTest,
    author = {Turing, A. M.},
    title = {I.—COMPUTING MACHINERY AND INTELLIGENCE},
    journal = {Mind},
    volume = {LIX},
    number = {236},
    pages = {433-460},
    year = {1950},
    month = {10},
    issn = {0026-4423},
    doi = {10.1093/mind/LIX.236.433},
    url = {https://doi.org/10.1093/mind/LIX.236.433},
    eprint = {https://academic.oup.com/mind/article-pdf/LIX/236/433/61209000/mind_lix_236_433.pdf},
}

@inproceedings{Levesque_WinogradSchemaChallenge,
author = {Levesque, Hector J. and Davis, Ernest and Morgenstern, Leora},
title = {The Winograd schema challenge},
year = {2012},
isbn = {9781577355601},
publisher = {AAAI Press},
booktitle = {Proceedings of the Thirteenth International Conference on Principles of Knowledge Representation and Reasoning},
pages = {552–561},
numpages = {10},
location = {Rome, Italy},
series = {KR'12}
}

@article{Nilsson_EmploymentTest,
author = {Nilsson, Nils J.},
title = {Human-Level Artificial Intelligence? Be Serious!},
journal = {AI Magazine},
volume = {26},
number = {4},
pages = {68-75},
doi = {https://doi.org/10.1609/aimag.v26i4.1850},
url = {https://onlinelibrary.wiley.com/doi/abs/10.1609/aimag.v26i4.1850},
eprint = {https://onlinelibrary.wiley.com/doi/pdf/10.1609/aimag.v26i4.1850},
year = {2005}
}

@book{russell_norvig_2010_AIModernApproach,
  title={Artificial Intelligence: A Modern Approach},
  author={Russell, Stuart J and Norvig, Peter},
  volume={3},
  year={2010},
  publisher={Prentice Hall},
  address={Upper Saddle River, NJ},
  isbn={978-0136042594},
  pages = {5}
}

@article{Searle_1980_sentience, title={Minds, brains, and programs}, volume={3}, DOI={10.1017/S0140525X00005756}, number={3}, journal={Behavioral and Brain Sciences}, author={Searle, John R.}, year={1980}, pages={417–424}}

@misc{chollet2019measureintelligence,
      title={On the Measure of Intelligence}, 
      author={François Chollet},
      year={2019},
      eprint={1911.01547},
      archivePrefix={arXiv},
      primaryClass={cs.AI},
      url={https://arxiv.org/abs/1911.01547}, 
}

@article{legg2007universal,
  title={Universal intelligence: A definition of machine intelligence},
  author={Legg, Shane and Hutter, Marcus},
  journal={Minds and machines},
  volume={17},
  number={4},
  pages={391--444},
  year={2007},
  publisher={Springer}
}

@misc{Hassabis2025XUniversalIntelligence,
  author       = {Hassabis, Demis},
  title        = {Yann is just plain incorrect here, he's confusing general intelligence with universal intelligence},
  year         = {2025},
  month        = dec,
  day          = {22},
  howpublished = {X (formerly Twitter) post},
  url          = {https://x.com/demishassabis/status/2003097405026193809},
  urldate      = {2026-01-26},
  note         = {Posted by @demishassabis}
}

@misc{Musk2025DemisIsRight,
  author       = {Musk, Elon},
  title        = {Demis is right},
  year         = {2025},
  month        = dec,
  day          = {22},
  howpublished = {X (formerly Twitter) post},
  url          = {https://x.com/elonmusk/status/2003165966243598738},
  urldate      = {2026-01-26},
  note         = {Posted by @elonmusk}
}

@ARTICLE{Wolpert_NoFreeLunch,
  author={Wolpert, D.H. and Macready, W.G.},
  journal={IEEE Transactions on Evolutionary Computation}, 
  title={No free lunch theorems for optimization}, 
  year={1997},
  volume={1},
  number={1},
  pages={67-82},
  doi={10.1109/4235.585893}}

@article{Wang2019_OnDefiningArtificialIntelligence,
author = {Wang, Pei},
year = {2019},
month = {08},
pages = {1-37},
title = {On Defining Artificial Intelligence},
volume = {10},
journal = {Journal of Artificial General Intelligence},
doi = {10.2478/jagi-2019-0002}
}

@inproceedings{Morris_LevelsOfAGI,
author = {Morris, Meredith Ringel and Sohl-Dickstein, Jascha and Fiedel, Noah and Warkentin, Tris and Dafoe, Allan and Faust, Aleksandra and Farabet, Clement and Legg, Shane},
title = {Position: levels of AGI for operationalizing progress on the path to AGI},
year = {2024},
publisher = {JMLR.org},
booktitle = {Proceedings of the 41st International Conference on Machine Learning},
articleno = {1478},
numpages = {14},
location = {Vienna, Austria},
series = {ICML'24}
}

@misc{hendrycks2025definitionagi,
      title={A Definition of AGI}, 
      author={Dan Hendrycks and Dawn Song and Christian Szegedy and Honglak Lee and Yarin Gal and Erik Brynjolfsson and Sharon Li and Andy Zou and Lionel Levine and Bo Han and Jie Fu and Ziwei Liu and Jinwoo Shin and Kimin Lee and Mantas Mazeika and Long Phan and George Ingebretsen and Adam Khoja and Cihang Xie and Olawale Salaudeen and Matthias Hein and Kevin Zhao and Alexander Pan and David Duvenaud and Bo Li and Steve Omohundro and Gabriel Alfour and Max Tegmark and Kevin McGrew and Gary Marcus and Jaan Tallinn and Eric Schmidt and Yoshua Bengio},
      year={2025},
      eprint={2510.18212},
      archivePrefix={arXiv},
      primaryClass={cs.AI},
      url={https://arxiv.org/abs/2510.18212}, 
}

@misc{openai_charter,
	title = {{OpenAI} {Charter}},
	url = {https://openai.com/charter/},
	language = {en-US},
	urldate = {2026-01-27},
    author = {OpenAI}
}

@article{
Mitchell2024_DebatesOnTheNatureOfAGI,
author = {Melanie Mitchell },
title = {Debates on the nature of artificial general intelligence},
journal = {Science},
volume = {383},
number = {6689},
pages = {eado7069},
year = {2024},
doi = {10.1126/science.ado7069},
URL = {https://www.science.org/doi/abs/10.1126/science.ado7069},
eprint = {https://www.science.org/doi/pdf/10.1126/science.ado7069}}

@article{lecun2022path,
  title={A path towards autonomous machine intelligence version 0.9. 2, 2022-06-27},
  author={LeCun, Yann},
  journal={Open Review},
  volume={62},
  number={1},
  pages={1--62},
  year={2022}
}

@book{craik1967nature,
  title={The nature of explanation},
  author={Craik, Kenneth James Williams},
  volume={445},
  year={1967},
  publisher={CUP Archive}
}

@inproceedings{he2020momentum,
  title={Momentum contrast for unsupervised visual representation learning},
  author={He, Kaiming and Fan, Haoqi and Wu, Yuxin and Xie, Saining and Girshick, Ross},
  booktitle={Proceedings of the IEEE/CVF conference on computer vision and pattern recognition},
  pages={9729--9738},
  year={2020}
}

@article{grill2020bootstrap,
  title={Bootstrap your own latent-a new approach to self-supervised learning},
  author={Grill, Jean-Bastien and Strub, Florian and Altch{\'e}, Florent and Tallec, Corentin and Richemond, Pierre and Buchatskaya, Elena and Doersch, Carl and Avila Pires, Bernardo and Guo, Zhaohan and Gheshlaghi Azar, Mohammad and others},
  journal={Advances in neural information processing systems},
  volume={33},
  pages={21271--21284},
  year={2020}
}

@inproceedings{chen2020simple,
  title={A simple framework for contrastive learning of visual representations},
  author={Chen, Ting and Kornblith, Simon and Norouzi, Mohammad and Hinton, Geoffrey},
  booktitle={International conference on machine learning},
  pages={1597--1607},
  year={2020},
  organization={PmLR}
}

@inproceedings{he2022masked,
  title={Masked autoencoders are scalable vision learners},
  author={He, Kaiming and Chen, Xinlei and Xie, Saining and Li, Yanghao and Doll{\'a}r, Piotr and Girshick, Ross},
  booktitle={Proceedings of the IEEE/CVF conference on computer vision and pattern recognition},
  pages={16000--16009},
  year={2022}
}

@inproceedings{lin2021limitations,
  title={Limitations of autoregressive models and their alternatives},
  author={Lin, Chu-Cheng and Jaech, Aaron and Li, Xin and Gormley, Matthew R and Eisner, Jason},
  booktitle={Proceedings of the 2021 conference of the North American chapter of the association for computational linguistics: Human language technologies},
  pages={5147--5173},
  year={2021}
}

@online{lecun_objective_driven_ai_harvard_2024,
  author  = {Yann LeCun},
  title   = {Objective-Driven AI: Towards AI systems that can learn, remember, reason, and plan},
  year    = {2024},
  date    = {2024-03-28},
  note    = {Harvard CMSA Ding Shum Lecture; slide includes \(P(\mathrm{correct})=(1-e)^n\) and “diverges exponentially”},
  url     = {https://cmsa.fas.harvard.edu/media/lecun-20240328-harvard_reduced.pdf},
  urldate = {2025-11-24}
}

@article{Forister_RevisitingEvoOfEcoSpec,
author = {Forister, M. L. and Dyer, L. A. and Singer, M. S. and Stireman III, J. O. and Lill, J. T.},
title = {Revisiting the evolution of ecological specialization, with emphasis on insect–plant interactions},
journal = {Ecology},
volume = {93},
number = {5},
pages = {981-991},
doi = {https://doi.org/10.1890/11-0650.1},
url = {https://esajournals.onlinelibrary.wiley.com/doi/abs/10.1890/11-0650.1},
eprint = {https://esajournals.onlinelibrary.wiley.com/doi/pdf/10.1890/11-0650.1},
year = {2012}
}

@article{Futuyma1988_EvoOfEcoSpec,
   author = "Futuyma, Douglas J. and Moreno, Gabriel",
   title = "THE EVOLUTION OF ECOLOGICAL SPECIALIZATION", 
   journal= "Annual Review of Ecology, Evolution, and Systematics",
   year = "1988",
   volume = "19",
   number = "Volume 19, 1988",
   pages = "207-233",
   doi = "https://doi.org/10.1146/annurev.es.19.110188.001231",
   url = "https://www.annualreviews.org/content/journals/10.1146/annurev.es.19.110188.001231",
   publisher = "Annual Reviews",
   issn = "1545-2069",
   type = "Journal Article",
  }

@misc{ruder2017overviewmultitasklearningdeep,
      title={An Overview of Multi-Task Learning in Deep Neural Networks}, 
      author={Sebastian Ruder},
      year={2017},
      eprint={1706.05098},
      archivePrefix={arXiv},
      primaryClass={cs.LG},
      url={https://arxiv.org/abs/1706.05098}, 
}

@article{narayanan2025ai,
  title={AI as normal technology},
  author={Narayanan, Arvind and Kapoor, Sayash},
  journal={Knight First Amendment Institute},
  year={2025}
}

@article{
Silver2018_AlphaGo,
author = {David Silver  and Thomas Hubert  and Julian Schrittwieser  and Ioannis Antonoglou  and Matthew Lai  and Arthur Guez  and Marc Lanctot  and Laurent Sifre  and Dharshan Kumaran  and Thore Graepel  and Timothy Lillicrap  and Karen Simonyan  and Demis Hassabis },
title = {A general reinforcement learning algorithm that masters chess, shogi, and Go through self-play},
journal = {Science},
volume = {362},
number = {6419},
pages = {1140-1144},
year = {2018},
doi = {10.1126/science.aar6404},
URL = {https://www.science.org/doi/abs/10.1126/science.aar6404},
eprint = {https://www.science.org/doi/pdf/10.1126/science.aar6404}}

@article{Hannan_Freeman_1977_EcologyOfOrganizations,
author = {Hannan, Michael T. and Freeman, John},
title = {The Population Ecology of Organizations},
journal = {American Journal of Sociology},
volume = {82},
number = {5},
pages = {929-964},
year = {1977},
doi = {10.1086/226424},

URL = { 
        https://doi.org/10.1086/226424
},
eprint = { 
        https://doi.org/10.1086/226424
}
}

@article{Loasby_1983_EvoTheoEconChange,
    author = {Loasby, Brian J.},
    title = {An Evolutionary Theory of Economic Change},
    journal = {The Economic Journal},
    volume = {93},
    number = {371},
    pages = {652-654},
    year = {1983},
    month = {09},
    issn = {0013-0133},
    doi = {10.2307/2232409},
    url = {https://doi.org/10.2307/2232409},
    eprint = {https://academic.oup.com/ej/article-pdf/93/371/652/27198325/ej0652.pdf},
}

@article{Fedus2022_MixtureOfExperts,
author = {Fedus, William and Zoph, Barret and Shazeer, Noam},
title = {Switch transformers: scaling to trillion parameter models with simple and efficient sparsity},
year = {2022},
issue_date = {January 2022},
publisher = {JMLR.org},
volume = {23},
number = {1},
issn = {1532-4435},
journal = {J. Mach. Learn. Res.},
month = jan,
articleno = {120},
numpages = {39}
}

@article{Bylander1994,
title = {The computational complexity of propositional STRIPS planning},
journal = {Artificial Intelligence},
volume = {69},
number = {1},
pages = {165-204},
year = {1994},
issn = {0004-3702},
doi = {https://doi.org/10.1016/0004-3702(94)90081-7},
url = {https://www.sciencedirect.com/science/article/pii/0004370294900817},
author = {Tom Bylander}
}

@article{LittmanGoldsmithMundhenk1998,
	title = {The {Computational} {Complexity} of {Probabilistic} {Planning}},
	volume = {9},
	issn = {1076-9757},
	url = {https://jair.org/index.php/jair/article/view/10208},
	doi = {10.1613/jair.505},
	urldate = {2026-01-29},
	journal = {Journal of Artificial Intelligence Research},
	author = {Littman, M. L. and Goldsmith, J. and Mundhenk, M.},
	month = aug,
	year = {1998},
	pages = {1--36},
}

@Article{Jumper2021,
author={Jumper, John
and Evans, Richard
and Pritzel, Alexander
and Green, Tim
and Figurnov, Michael
and Ronneberger, Olaf
and Tunyasuvunakool, Kathryn
and Bates, Russ
and {\v{Z}}{\'i}dek, Augustin
and Potapenko, Anna
and Bridgland, Alex
and Meyer, Clemens
and Kohl, Simon A. A.
and Ballard, Andrew J.
and Cowie, Andrew
and Romera-Paredes, Bernardino
and Nikolov, Stanislav
and Jain, Rishub
and Adler, Jonas
and Back, Trevor
and Petersen, Stig
and Reiman, David
and Clancy, Ellen
and Zielinski, Michal
and Steinegger, Martin
and Pacholska, Michalina
and Berghammer, Tamas
and Bodenstein, Sebastian
and Silver, David
and Vinyals, Oriol
and Senior, Andrew W.
and Kavukcuoglu, Koray
and Kohli, Pushmeet
and Hassabis, Demis},
title={Highly accurate protein structure prediction with AlphaFold},
journal={Nature},
year={2021},
month={Aug},
day={01},
volume={596},
number={7873},
pages={583-589},
issn={1476-4687},
doi={10.1038/s41586-021-03819-2},
url={https://doi.org/10.1038/s41586-021-03819-2}
}

@article{TverskyKahneman1974,
author = {Amos Tversky  and Daniel Kahneman },
title = {Judgment under Uncertainty: Heuristics and Biases},
journal = {Science},
volume = {185},
number = {4157},
pages = {1124-1131},
year = {1974},
doi = {10.1126/science.185.4157.1124},
URL = {https://www.science.org/doi/abs/10.1126/science.185.4157.1124},
eprint = {https://www.science.org/doi/pdf/10.1126/science.185.4157.1124}}

@article{Li2018Mismatch,
author = {Norman P. Li and Mark van Vugt and Stephen M. Colarelli},
title ={The Evolutionary Mismatch Hypothesis: Implications for Psychological Science},
journal = {Current Directions in Psychological Science},
volume = {27},
number = {1},
pages = {38-44},
year = {2018},
doi = {10.1177/0963721417731378},
URL = {https://doi.org/10.1177/0963721417731378},
eprint = {https://doi.org/10.1177/0963721417731378}
}

@article{Domingos2012,
author = {Domingos, Pedro},
title = {A few useful things to know about machine learning},
year = {2012},
issue_date = {October 2012},
publisher = {Association for Computing Machinery},
address = {New York, NY, USA},
volume = {55},
number = {10},
issn = {0001-0782},
url = {https://doi.org/10.1145/2347736.2347755},
doi = {10.1145/2347736.2347755},
journal = {Commun. ACM},
month = oct,
pages = {78–87},
numpages = {10}
}

@misc{zhao2025absolutezeroreinforcedselfplay,
      title={Absolute Zero: Reinforced Self-play Reasoning with Zero Data}, 
      author={Andrew Zhao and Yiran Wu and Yang Yue and Tong Wu and Quentin Xu and Yang Yue and Matthieu Lin and Shenzhi Wang and Qingyun Wu and Zilong Zheng and Gao Huang},
      year={2025},
      eprint={2505.03335},
      archivePrefix={arXiv},
      primaryClass={cs.LG},
      url={https://arxiv.org/abs/2505.03335}, 
}

@Article{LeCun2015DeepLearning,
author={LeCun, Yann
and Bengio, Yoshua
and Hinton, Geoffrey},
title={Deep learning},
journal={Nature},
year={2015},
month={May},
day={01},
volume={521},
number={7553},
pages={436-444},
issn={1476-4687},
doi={10.1038/nature14539},
url={https://doi.org/10.1038/nature14539}
}

@INPROCEEDINGS{Assran2023_I_JEPA,
  author={Assran, Mahmoud and Duval, Quentin and Misra, Ishan and Bojanowski, Piotr and Vincent, Pascal and Rabbat, Michael and LeCun, Yann and Ballas, Nicolas},
  booktitle={2023 IEEE/CVF Conference on Computer Vision and Pattern Recognition (CVPR)}, 
  title={Self-Supervised Learning from Images with a Joint-Embedding Predictive Architecture}, 
  year={2023},
  volume={},
  number={},
  pages={15619-15629},
  doi={10.1109/CVPR52729.2023.01499}}

@misc{hafner2025_Dreamer4,
      title={Training Agents Inside of Scalable World Models}, 
      author={Danijar Hafner and Wilson Yan and Timothy Lillicrap},
      year={2025},
      eprint={2509.24527},
      archivePrefix={arXiv},
      primaryClass={cs.AI},
      url={https://arxiv.org/abs/2509.24527}, 
}

@misc{bruce2024_genie_generativeinteractiveenvironments,
      title={Genie: Generative Interactive Environments}, 
      author={Jake Bruce and Michael Dennis and Ashley Edwards and Jack Parker-Holder and Yuge Shi and Edward Hughes and Matthew Lai and Aditi Mavalankar and Richie Steigerwald and Chris Apps and Yusuf Aytar and Sarah Bechtle and Feryal Behbahani and Stephanie Chan and Nicolas Heess and Lucy Gonzalez and Simon Osindero and Sherjil Ozair and Scott Reed and Jingwei Zhang and Konrad Zolna and Jeff Clune and Nando de Freitas and Satinder Singh and Tim Rocktäschel},
      year={2024},
      eprint={2402.15391},
      archivePrefix={arXiv},
      primaryClass={cs.LG},
      url={https://arxiv.org/abs/2402.15391}, 
}

@misc{su2025largelanguagemodelsreally,
      title={Do Large Language Models (Really) Need Statistical Foundations?}, 
      author={Weijie Su},
      year={2025},
      eprint={2505.19145},
      archivePrefix={arXiv},
      primaryClass={stat.ME},
      url={https://arxiv.org/abs/2505.19145}, 
}

@Article{Huang2024_LLM2LMM,
AUTHOR = {Huang, Dawei and Yan, Chuan and Li, Qing and Peng, Xiaojiang},
TITLE = {From Large Language Models to Large Multimodal Models: A Literature Review},
JOURNAL = {Applied Sciences},
VOLUME = {14},
YEAR = {2024},
NUMBER = {12},
ARTICLE-NUMBER = {5068},
URL = {https://www.mdpi.com/2076-3417/14/12/5068},
ISSN = {2076-3417},
DOI = {10.3390/app14125068}
}

@article{wyder2025common,
  title={Common task framework for a critical evaluation of scientific machine learning algorithms},
  author={Wyder, Philippe Martin and Goldfeder, Judah and Yermakov, Alexey and Zhao, Yue and Riva, Stefano and Williams, Jan P and Zoro, David and Rude, Amy Sara and Tomasetto, Matteo and Germany, Joe and others},
  journal={arXiv preprint arXiv:2510.23166},
  year={2025}
}

@misc{balestriero2023cookbookselfsupervisedlearning,
      title={A Cookbook of Self-Supervised Learning}, 
      author={Randall Balestriero and Mark Ibrahim and Vlad Sobal and Ari Morcos and Shashank Shekhar and Tom Goldstein and Florian Bordes and Adrien Bardes and Gregoire Mialon and Yuandong Tian and Avi Schwarzschild and Andrew Gordon Wilson and Jonas Geiping and Quentin Garrido and Pierre Fernandez and Amir Bar and Hamed Pirsiavash and Yann LeCun and Micah Goldblum},
      year={2023},
      eprint={2304.12210},
      archivePrefix={arXiv},
      primaryClass={cs.LG},
      url={https://arxiv.org/abs/2304.12210}, 
}

@article{sutton2019bitter,
  title={The bitter lesson},
  author={Sutton, Richard},
  journal={Incomplete Ideas (blog)},
  volume={13},
  number={1},
  pages={38},
  year={2019}
}

@article{van2025joint,
  title={Joint Embedding vs Reconstruction: Provable Benefits of Latent Space Prediction for Self Supervised Learning},
  author={Van Assel, Hugues and Ibrahim, Mark and Biancalani, Tommaso and Regev, Aviv and Balestriero, Randall},
  journal={arXiv preprint arXiv:2505.12477},
  year={2025}
}

@inproceedings{finn2017model,
  title={Model-agnostic meta-learning for fast adaptation of deep networks},
  author={Finn, Chelsea and Abbeel, Pieter and Levine, Sergey},
  booktitle={International conference on machine learning},
  pages={1126--1135},
  year={2017},
  organization={PMLR}
}

@Article{Schrittwieser2020,
author={Schrittwieser, Julian
and Antonoglou, Ioannis
and Hubert, Thomas
and Simonyan, Karen
and Sifre, Laurent
and Schmitt, Simon
and Guez, Arthur
and Lockhart, Edward
and Hassabis, Demis
and Graepel, Thore
and Lillicrap, Timothy
and Silver, David},
title={Mastering Atari, Go, chess and shogi by planning with a learned model},
journal={Nature},
year={2020},
month={Dec},
day={01},
volume={588},
number={7839},
pages={604-609},
issn={1476-4687},
doi={10.1038/s41586-020-03051-4},
url={https://doi.org/10.1038/s41586-020-03051-4}
}

@article{aguerra_y_arcas_norvig_2023,
  author       = {Agüera y Arcas, Blaise and Norvig, Peter},
  title        = {Artificial General Intelligence Is Already Here},
  journal      = {Noema Magazine},
  year         = {2023},
  month        = {October},
  url          = {https://www.noemamag.com/artificial-general-intelligence-is-already-here/},
  note         = {Accessed: YYYY-MM-DD}
}

@article{chen_belkin_bergen_danks_2026,
  author       = {Chen, Eddy Keming and Belkin, Mikhail and Bergen, Leon and Danks, David},
  title        = {Does AI already have human-level intelligence? The evidence is clear},
  journal      = {Nature},
  year         = {2026},
  volume       = {650},
  pages        = {36--40},
  doi          = {10.1038/d41586-026-00285-6},
  url          = {https://www.nature.com/articles/d41586-026-00285-6},
  note         = {Accessed: YYYY-MM-DD}
}
\bibliographystyle{icml2026}

\newpage
\appendix
\onecolumn

\end{document}